\documentclass{article}
\usepackage{ijcai22}

\usepackage{times}
\usepackage{soul}
\usepackage{url}
\usepackage[hidelinks]{hyperref}
\usepackage[utf8]{inputenc}
\usepackage[small]{caption}
\usepackage{graphicx}
\usepackage{amsmath}
\usepackage{amsthm}
\usepackage{booktabs}
\usepackage{algorithm}
\usepackage{algorithmic}
\usepackage{amsmath}
\usepackage{amssymb}
\usepackage{dsfont}
\usepackage{multirow}
\title{Fake Motion Video Generation: Copy Motion From One to Another}
\title{Copy Motion From One to Another: Fake Motion Video Generation}

\author{
Zhenguang Liu$^{1,2}$\and
Sifan Wu$^2$\footnotemark[1]\and
Chejian Xu$^1$\footnote{Corresponding Authors }\and
Xiang Wang$^3$\and
Lei Zhu$^4$\and
Shuang Wu$^5$\And
Fuli Feng$^6$\\
\affiliations
$^1$Zhejiang University\\
$^2$Zhejiang Gongshang University\\
$^3$National University of Singapore\\
$^4$Shandong Normal Unversity\\
$^5$Nanyang Technological University\\
$^6$University of Science and Technology of China\\
\emails
\{liuzhenguang2008,wusifan2021\}@gmail.com,
xuchejian@zju.edu.cn,
xiangwang@u.nus.edu,
leizhu0608@gmail.com,
wushuang@outlook.sg,
fulifeng93@gmail.com
}

\begin{document}

\maketitle

\begin{abstract}
One compelling application of artificial intelligence is to generate a video of a target person performing arbitrary desired motion (from a source person). While the state-of-the-art methods are able to synthesize a video demonstrating similar broad stroke motion details, they are generally lacking in texture details. A pertinent manifestation appears as distorted \emph{face,  feet}, and \emph{hands}, and such flaws are very sensitively perceived by human observers. Furthermore, current methods typically employ GANs with a L2 loss to assess the authenticity of the generated videos, inherently requiring a large amount of training samples to learn the texture details for adequate video generation. In this work, we tackle these challenges from three aspects: 1) We disentangle each video frame into foreground (the person) and background, focusing on generating the foreground to reduce the underlying dimension of the network output. 2) We propose a theoretically motivated Gromov-Wasserstein loss that facilitates learning the mapping from a pose to a foreground image. 3) To enhance texture details, we encode facial features with geometric guidance and employ local GANs to refine the face, feet, and hands. Extensive experiments show that our method is able to generate realistic target person videos, faithfully copying complex motions from a source person. 
	
\end{abstract}

\section{Introduction}
	The recent advancement in artificial intelligence has taken the world by storm and given rise to a lot of interesting and compelling applications \cite{liu2019towards,liu2021motion,song2021spatial,wang2021visual}. Motion copy is one emerging research problem which looks at generating a fake video of a target person performing arbitrary motion, usually extracted from a source person. While the two person may be vastly different in body shape and appearance, motion copy allows retargeting the motion to the target, empowering an untrained person to be depicted in videos dancing like pop stars \cite{chan2019everybody}, playing like NBA players, and maneuvers in gymnastics and martial arts.

	Fundamentally, copying the motion from a source person to a target person amounts to learning a mapping between their video frames. Due to the high dimensionality of this mapping (which renders it almost intractable), it would be more economical to leverage an intermediate representation to bridge the problem. Typically, such approaches include pose-guided and mesh-guided representations, which respectively extract pose or mesh representations from the source person and then learn a generative model that maps this intermediate representation (pose or mesh) to the appearance of the target person. We focus on pose-guided target video generation in this paper.


	Upon scrutinizing and experimenting with the released implementations of the state-of-the-art methods, we observe that there are \textbf{two key issues} to be addressed. \textbf{(1)} Whereas existing methods achieve plausible results on a broad stroke, the issues of distorted \emph{face}, \emph{hands}, and \emph{feet} are quite rampant and such defects are highly noticeable to the human observer. \textbf{(2)} Current approaches tend to heavily rely on GANs with L2 loss to learn generative models for pose-to-appearance generation. When the training samples for the target person are scarce (typically only a single target video is available), it would be inadequate to learn a competent GAN that typically requires a large amount of training samples to discriminate the authenticity of the generated frames satisfactorily.

To tackle the challenges, we propose a novel motion copy framework \emph{FakeMotion}, consisting of three key designs: 1) We first perform a foreground and background separation. Focusing on foreground generation fundamentally reduces the complexity and dimensionality of the problem. 2) We propose a theoretically motivated Gromov-Wasserstein loss to guide pose-to-appearance generation. The Gromov-Wasserstein loss constitutes an optimal transport objective that learns the pose-to-appearance mapping in a pair-wise fashion, and is more data-efficient and well-suited than conventional GANs for learning the generative model. 3) We refine the textural details of the \emph{face, feet, and hands} with multiple dedicated GANs. In particular, our face refinement adopts a self-supervised approach where we employ geometric cues such as facial orientations to serve as guidance for mining real facial appearances of the target person. These are combined with the coarse generated face for refinement, and the encoding of similar oriented real faces results in more delicate and realistic facial textures. Extensive experiments demonstrate that our method outperforms existing methods.


To summarize, the \textbf{key contributions} of this work are: 	\textbf{(1)} We propose a novel architecture for fake motion video generation, which exploits a Gromov-Wasserstein loss to frame pairwise distance constraint in pose-to-appearance generation. We also present a dense skip connection structure in the generator to facilitate filling the gap between pose and appearance. \textbf{(2)} We further introduce a local enhancement module to explicitly handle the \emph{face}, \emph{feet}, and \emph{hands}, which are visually prominent body parts and are often affected by artifacts. \textbf{(3)} Extensive experiments on benchmark datasets demonstrate that our approach achieves new state-of-the-art performance. As a side contribution, we crafted a new dataset for evaluating motion copy methods where 50 subjects in the videos perform complex motions.  

\begin{figure*}[ht]
	\centering
	\includegraphics[width=15.6cm]{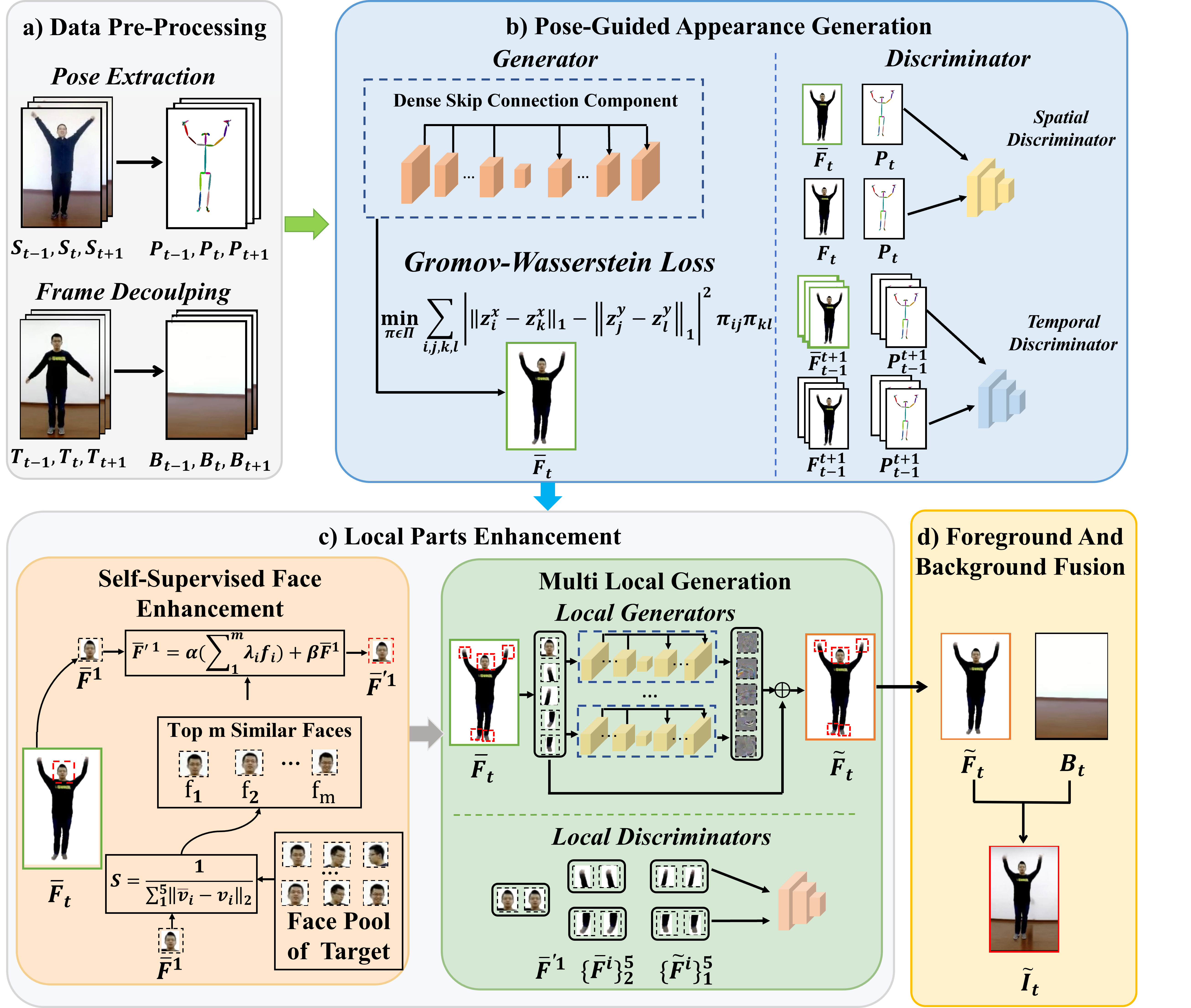}
	\caption{\textbf{Overall pipeline.} a) For \emph{data pre-processing}, we extract poses $\{P_i\}_{i=t-1}^{t+1}$ from source video frames $\{S_i\}_{i=t-1}^{t+1}$ and decouple the target video frames $\{T_i\}_{i=t-1}^{t+1}$ into background and foreground. b) The \emph{pose-guided appearance generation} module synthesizes the corresponding foreground image of pose $P_t$ given a conditioned input of $\{P_i\}_{i=t-1}^{t+1}$. We employ a Gromov-Wasserstein loss to facilitate matching of the generated image features with the real image features. Our discriminator assesses spatial-temporal consistency in the video. c) The \emph{local parts enhancement} module refines the details for the \emph{faces, hands and feet}. d) Finally, we \emph{fuse} foreground with the background to generate the entire frame.}
	
	\label{fig_overview}
\end{figure*}

\section{Method}
\label{headings}

\paragraph{Problem Statement.} Given two videos, one for the target person whose appearance we want to synthesize, and the other for the source person whose motion we seek to copy \cite{chan2019everybody}, the goal is to synthesize a video of the target person enacting the same motion as the source person. 

\paragraph{Method Overview.} An overview of our method \emph{Fakemotion} is outlined in Fig.~\ref{fig_overview}. Overall, \emph{Fakemotion} consists of three key stages: a) {Data Pre-processing},  b) {Pose-Guided Appearance Generation}, and c) {Local Parts Enhancement}. Specifically, we \emph{first} extract the pose sequence from the video of the source person. \emph{Then}, we feed the poses into a pose-to-appearance network to synthesize the foreground images of the target person. \emph{Further}, we engage in multiple local GANs to polish the local body parts. \emph{Finally}, the polished foreground is fused with the background, which is extracted from the video of the target person, to approach the final result. We would like to highlight that our generators  have an edge in adopting dense skip connections and are equipped with a Gromov Wasserstein loss, while our discriminator games in spatial and temporal dual constraints. In what follows, we elaborate on the key components in detail.

\subsection{Data Pre-processing}
Presented with a video $V_s$ of the source person and a video $V_t$ of the target person, we perform the following in parallel: (1) extracting the human pose from each frame of $V_s$, and (2) disentangling each frame of $V_t$ into foreground (person) $F$ and background $B$. The motivations are twofold: (1) The extracted poses of source $V_s$ unambiguously characterize the motion and can be used to guide appearance generation.  (2) Generating the entire image conditioned on a pose might be too ambitious. Synthesizing only the foreground conditioned on a pose input would reduce the complexity and result in better texture details. Technically, we could adopt off-the-shelf pose estimators \cite{cao2019openpose,liu2021deep,liu2022temporal} for pose extraction.
For foreground and background decoupling, we follow \cite{he2017mask}.

\subsection{Pose-to-appearance Generation} 
Now, our goal is to generate an appealing foreground image in accordance with a given pose.  Particularly, we design a pose-to-appearance generation network, consisting of a \emph{generator} that incorporates dense skip connections and a Gromov-Wasserstein loss, 
as well as a \emph{discriminator} that enforces spatio-temporal dual constraints. 

\begin{figure*}[ht]
	\centering
	\includegraphics[width=18cm]{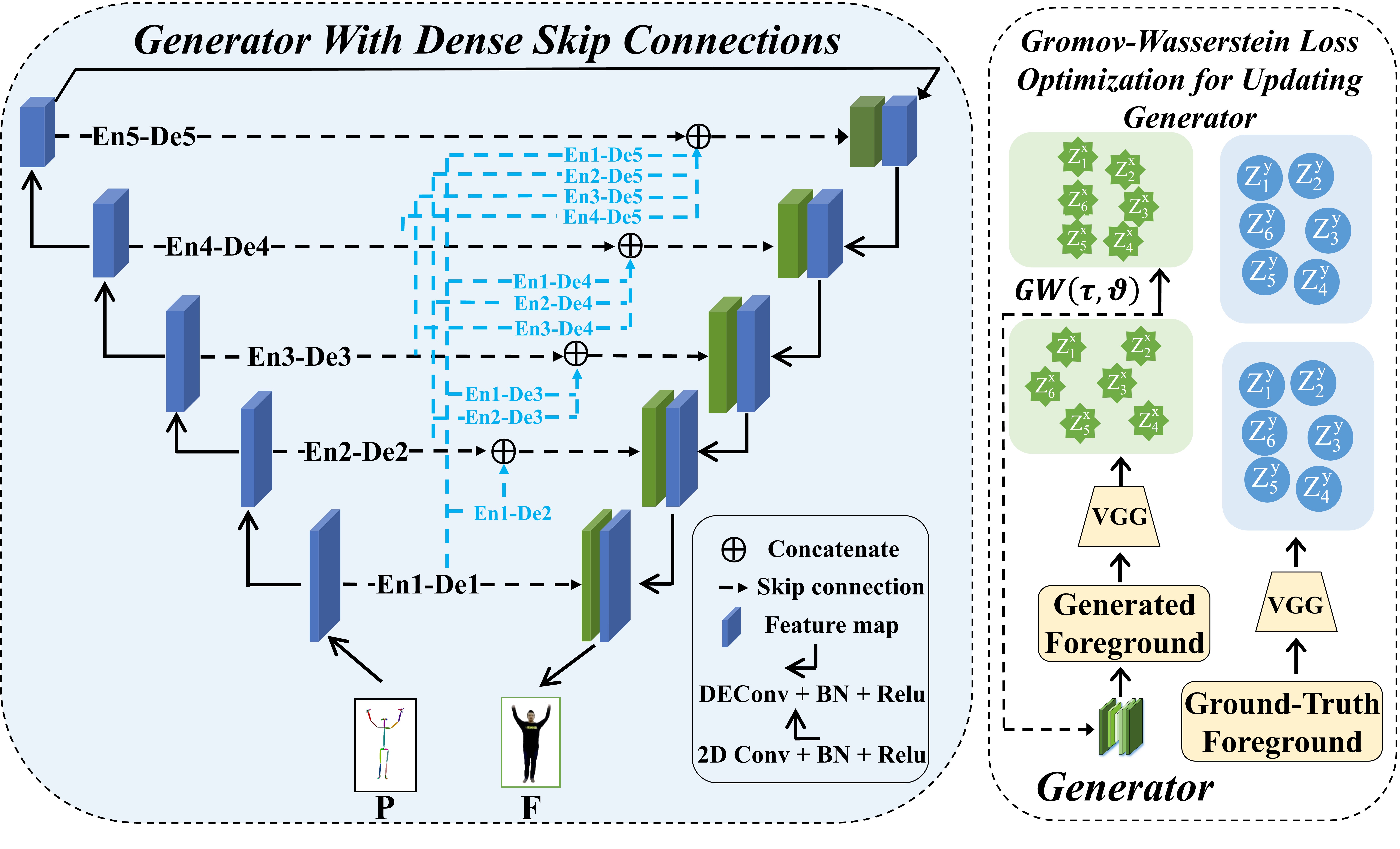}
	\caption{\textbf{Generator and Gromov-Wasserstein Loss.} Left: An encoder-decoder architecture with dense skip connections that facilitate the fusion of features across different scales. The blue lines represent the extra connections we added. Right: We introduce the Gromov-Wasserstein loss to update the parameters of the generator and minimize the discrepancy between intra-space distances.}
	\label{dense}
\end{figure*}

\paragraph{Generator with Dense Skip Connections.}
As illustrated in Fig.~\ref{dense}, we engage a U-shaped architecture for our pose-to-appearance network. 
In vanilla U-Net, a decoder layer $D_i$ is connected to a single encoder layer $E_i$ and is unable to access low-level features from encoders $\{E_j\}_{j=1}^{i-1}$. Consequently, different level features are isolated to each other and their connections are ignored. Building upon the U-Net, we propose to add extra skip connections that help a decoder layer to access lower level features. Instead of connecting a decoder $D_i$ at layer $i$ with only the encoder $E_i$ at layer $i$, we allow $D_i$ to access encoder layers $\big\{E_j|j<i\}$. In this way, each decoder could integrate and fuse feature representations from multiple layers.

\paragraph{Gromov-Wasserstein Loss to Facilitate Pose-to-appearance Generation.} Our generator network is designed to learn a mapping from a pose $P_t$ to the corresponding foreground image $F_t$ of the target person. To enhance temporal consistency, instead of a single pose $P_t$, we input a pose context window $\langle P_{t-1}, P_t, P_{t+1} \rangle$ to the generator. Current methods typically approach this pose-to-appearance problem with a conventional GAN, and measure the discrepancy between the generated foreground and ground truth via a pixelwise L2 loss. However, this inherently requires a large number of training samples in order to generate viable and realistic images. Furthermore, training such GAN models is often precarious and tends to be unstable when adapted to a wide range of target videos. To alleviate this issue, inspired by optimal transport, we propose to measure the generated and ground truth distance via a Gromov-Wasserstein loss, which essentially defines a pairwise distance at the feature-level.

Interestingly, our experiments reveal that our pairwise loss plays an unexpectedly significant role in the model performance. Heuristically, our proposed Gromov-Wasserstein loss looks at the overall distance structure perception. We first extract the VGG features for the generated foreground $\overline{F}$ and ground truth $F$:

\begin{align}
	\begin{split}
		z^x = \Psi(\overline{F}), z^y = \Psi(F),
	\end{split}
\end{align}
where $\Psi(\cdot) $ represents a pre-trained VGG network. The Gromov-Wasserstein loss defines a distance between the metrics on the two feature spaces. We consider two batches of feature vectors $\{z^x_i\}^n_{i=1}$, $\{z^y_i\}^n_{i=1}$ as two discrete empirical distributions $\tau$, $\upsilon$:

\begin{align}
	\begin{split}
		\tau = \sum_{i=1}^{n}\frac{1}{n}\delta_{z^x_i}, \upsilon = \sum_{i=1}^{n}\frac{1}{n}\delta_{z^y_i},
	\end{split}
\end{align}
where $\delta$ denotes the Dirac delta distribution. Formally, the Gromov-Wasserstein loss for our task is defined as

\begin{align}
	GW(\tau,\upsilon) = \min_{\pi\in \Pi }\sum _{i,j,k,l}\left | \left \| z^x_i - z^x_k  \right \|_1 - \left \| z^y_j - z^y_l  \right \|_1 \right |^2 \pi _{ij}\pi _{kl},
\end{align}
where $\Pi$ denotes the set of point distributions with marginals $\tau$ and $\upsilon$. To obtain the Gromov-Wasserstein distance for these two point distributions, we solve for the optimal transport matrix $\pi$ that minimizes the squared distance between the intra-space L1 costs. We follow \cite{peyre2016gromov} in introducing an entropic regularization term which guarantees tractability (as well as being amenable to backpropagation) in the optimization. We then adopt the Sinkhorn algorithm and projected gradient descent to optimize the entropy regularized Gromov-Wasserstein objectives.

\begin{align}
	&C = \left | \left \| z^x_i - z^x_k  \right \|_1 - \left \| z^y_j - z^y_l  \right \|_1 \right |^2, \\
	&K=exp(-C/\varepsilon),\\
	&a\leftarrow \mathds{1}_n/Kb, b\leftarrow \mathds{1}_n/K^Ta,\\
	&\pi = diag(a)Kdiag(b),
\end{align}
where C denotes the cost matrix for a given batch of the generated foreground features and the ground-truth features. $\varepsilon$ denotes a regularization coefficient.

Optimizing the Gromov-Wasserstein distance via above equations (Eqs.4--7) updates the generator network parameters through backpropagation. Effectively, this effectuates an alignment of these two feature embedding spaces that facilitates matching of the generated foreground image with the ground-truth. Even when there is scarce training samples for the target person, this pairwise feature matching approach could train the network sufficiently and yield a more detailed and realistic generation. Furthermore, the Gromov-Wasserstein distance is also tractable and stable, and empirically generalizes well to a variety of target video domains.

\paragraph{Discriminator with Dual Constraints.} Humans generally judge the authenticity of a video from two aspects: the quality of the images and the temporal consistency across frames. The continuity between frames plays a crucial role in visual judgment. Yet, this is often neglected in many existing methods \cite{dong2018soft,kappel2021high} which tend to focus on single frame quality and cannot explicitly consider temporal continuity. In our approach, we devise dual constraints which are composed of a quality discriminator $D_q$ and a temporal discriminator $D_t$. (1) The quality discriminator $D_q$ focuses on the single frame quality to assess the authenticity of the forged foreground. (2) The temporal discriminator $D_t$ is equipped with a set of parallel dilated convolutions that capture temporal information across frames at various temporal scales. As illustrated in Fig.~\ref{fig_overview}, $D_q$ takes $\left ( P_t,{F}_t \right )$ and $( P_t,\overline{F}_{t})$
as input pairs while the input to $D_t$ consists of  $(\{P_i\}_{i=t-1}^{t+1}, \{F_i\}_{i=t-1}^{t+1}) $ and $(\{P_i\}_{i=t-1}^{t+1}, \{ \overline{F_i}\}_{i=t-1}^{t+1}) $. Both discriminators are trained to classify the authenticity of the generated frame or context window.

\subsection{Local Parts Enhancement}
While the global quality of the generated frames is certainly essential when judging the authenticity of a video, our visual system is also sensitive to prominent body parts, especially the face as well as the hands and feet. When these body parts appear unnatural, our mind would be quick to identify them and judge the videos as fake. Upon experimenting with existing methods following their released codes \cite{chan2019everybody,balakrishnan2018synthesizing,wang2019few,liu2019liquid,wei2020c2f}, an important insight and conclusion that we draw are that even the state-of-the-art methods still have difficulties in generating {natural} and realistic details for the face, hands, and feet. Therefore, we introduce a self-supervised face enhancement and additional dedicated local GANs to refine these  body parts, with the goal of polishing the texture details on top of the generated results from the pose-to-appearance generation module. 


\begin{figure}[ht]
	\centering
	\includegraphics[width=8.5cm]{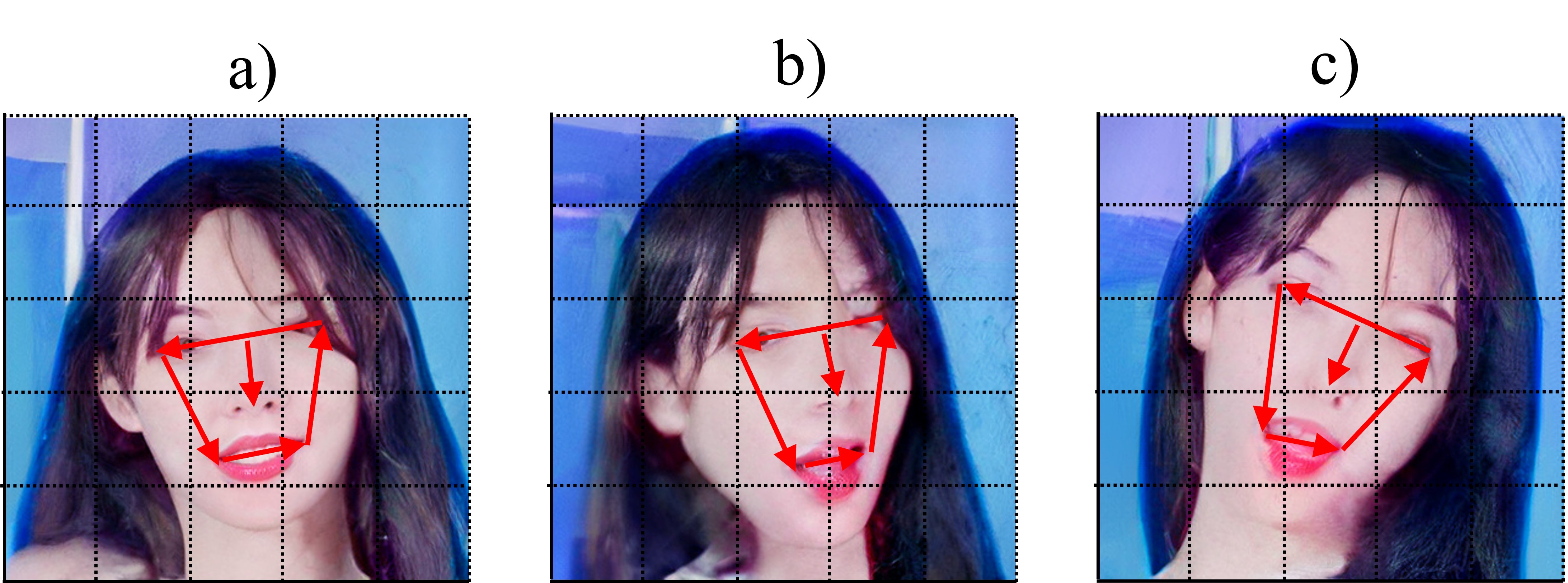}
	\caption{\textbf{Face Orientation} is extracted from the face vector field. Three different face orientations are presented.} 
	\label{face}	
\end{figure}

\paragraph{Self-Supervised Face Enhancement.} Intuitively, the facial images of the same person should look alike when they are viewed from the similar orientation or viewing angles. With this geometric guidance, we search from the given  video of the target person to extract facial images that exhibit similar facial orientations and alignment as that of the synthesized image. This is then leveraged to refine the final face output.

To measure face orientation similarity, we do not simply compute the (cosine) similarity between facial features as the full facial features would usually contain redundant information, \emph{e.g., color and eye shape}, that are irrelevant to face orientation. Instead, we represent facial orientation with vector fields. Specifically, as shown in Fig.~3, we employ 5 vectors, including $v_{1}$: right eye $\rightarrow$ left eye, $v_2$: left eye $\rightarrow$ left side of the mouth, $v_3$: left side of the mouth $\rightarrow$ right side of the mouth, $v_4$: right side of the mouth $\rightarrow$ right eye, $v_5$: the center of eyes $\rightarrow$ nose. Given two face orientations $\{v_i\}_{i=1}^5$ and $ \{\overline{v}_i\}_{i=1}^5$, their similarity can be conveniently computed as:
\begin{align}
	\mathcal{S}=\frac{1} {\sum_{k=1}^5\|\overline{v}_i-v_i\|_2},
\end{align}

Subsequently, we select top $m$ ground-truth face images $\mathbf{f}$ = $\lbrace${$f_1$ ,..., $f_m$}$\rbrace$  with the highest similarities as 
auxiliary faces. Finally, the initially generated face $\overline{F}^{\prime 1}$ is enhanced into:
\begin{align}
	\overline{F}^{\prime 1} = \alpha (\sum_{i=1}^m( \lambda_i  f_i)+\beta \overline{F}^{1},
\end{align}
where $\alpha$, $\lambda_i$, and $\beta$  are hyperparameters.
\par



\paragraph{Multi-Local GANs for body parts.} The face enhancement enforces the generated face to be similar to faces with close orientations. We further polish the face and limbs. Similar to the divide-and-conquer strategy in \cite{liu2021aggregated}, we model each body part with a dedicated local GAN to perform refinement. Specifically, we crop the five key parts $\overline{{F}}^{\prime 1}$ (face) and $\{\overline{{F}}^{i}\}_2^5$ (hands, and feet) of the generated foreground image $\overline{{F}}_{t}$, where $i$ denotes the $i^{th}$ key body part. We further feed them into corresponding GANs, which are trained to output a residual image ${F}_r^i$. The residual image reveals the difference in color and texture details between the generated image of the body part and the ground truth. We add them to the generated images to yield the final foreground:
\begin{align}
	&\widetilde{{F}}^1 = \overline{{F}}^{\prime 1} + {F}_r^1, \\
	&\widetilde{{F}}^i = \overline{{F}}^{i} + {F}_r^i.
\end{align}

\paragraph{Foreground and Background Fusion.} 
Finally we may combine the polished foreground $\widetilde{{F}}_t$ with the inpainted background ${B}_t$. We employ a linear sum to fuse $\widetilde{{F}}_t$ and ${B}_t$ into the final video frame as:
\begin{align}		
	\tilde{I_t} = M \odot \widetilde{{F}}_t + (1-M) \odot {B}_t
\end{align}
where $M$ is the mask obtained during the pre-processing phase highlighting the region of the foreground.

In practice, for background ${B}_t$, we filled the removed foreground pixels in ${B}_t$ following \cite{yu2019free}.

\subsection{Loss Functions}
Now, we zoom in on the loss functions. The proposed Gromov-Wasserstein loss 
promotes a better learning of the embedding correspondence between the generated foreground space and the ground-truth space. This facilitates extracting  commonalities and shared structural similarities between the two spaces. 

We further introduce two standard spatial-temporal adversarial losses for $D_q$ and $D_t$, where $D_q$ enhances the frame-by-frame spatial constraints and $D_t$ reinforces temporal constraints for every three consecutive frames:

\begin{align}
	\begin{split}
		L^{spatial} &= \mathbb{E}_{({p, I})}[\log D_q(P_t, {F}_t)]\\ 
		&+ \mathbb{E}_p[\log(1-D_q(P_t, G(P_t)))]\\
	\end{split}\\
	\begin{split}
	L^{temporal} &= \mathbb{E}_{({p, I})}[\log D_t(P_{t-1}^{t+1}, {F}_{t-1}^{t+1})]\\
	& + \mathbb{E}_p[\log(1-D_t(P_{t-1}^{t+1},G(P_{t-1}^{t+1})],		
	\end{split}
\end{align}

\begin{table*}[ht]\small
	\centering
	
	\label{compare}
	\begin{tabular}{c|ccc|ccc|ccc|ccc} 
		\hline
		\multirow{3}{*}{\textbf{Methods}} & \multicolumn{6}{c|}{\textbf{ComplexMotion}}                                                                                                                                                                   & \multicolumn{6}{c}{\textbf{iPER}}                                                                                                                                                                              \\ 
		\cline{2-13}
		& \multicolumn{3}{c|}{Person Reconstruction}                                                    & \multicolumn{3}{c|}{Action Imitation}                                                                         & \multicolumn{3}{c|}{Person Reconstruction}                                                    & \multicolumn{3}{c}{Action Imitation}                                                                           \\ 
		\cline{2-13}
		& \multicolumn{1}{c|}{SSIM$\uparrow$} & \multicolumn{1}{c|}{PSNR$\uparrow$} & LPIPS$\downarrow$ & \multicolumn{1}{c|}{FID$\downarrow$} & \multicolumn{1}{c|}{IS$\uparrow$} & \multicolumn{1}{l|}{TCM$\uparrow$} & \multicolumn{1}{c|}{SSIM$\uparrow$} & \multicolumn{1}{c|}{PSNR$\uparrow$} & LPIPS$\downarrow$ & \multicolumn{1}{c|}{FID$\downarrow$} & \multicolumn{1}{c|}{IS$\uparrow$} & \multicolumn{1}{l}{TCM $\uparrow$}  \\ 
		\hline
		EDN                               & 0.823                               & 24.36                               & 0.061             & 64.12                                & 3.411                             & 0.534                              & 0.852                               & 24.48                               & 0.086             & 57.52                                & 3.305                             & 0.591                               \\
		FSV2V                             & 0.748                               & 22.51                               & 0.132             & 99.11                                & 3.164                             & 0.575                              & 0.824                               & 21.18                               & 0.108             & 107.29                               & 3.136                             & 0.754                               \\
		PoseWarp                          & 0.711                               & 21.42                               & 0.149             & 78.21                                & 3.109                             & 0.334                              & 0.792                               & 22.16                               & 0.119             & 115.23                               & 3.095                             & 0.601                               \\
		LWGAN                             & 0.789                               & 24.27                               & 0.081             & 85.30                                & 3.398                             & 0.683                              & 0.843                               & 22.32                               & 0.091             & 76.38                                & 3.258                             & 0.729                               \\
		C2F-FWN                           & 0.878                               & 25.68                               & 0.048             & 53.19                                & 3.408                             & 0.689                              & 0.847                               & 24.32                               & 0.074             & 60.12                                & 3.412                             & 0.769                               \\ 
		\hline
		SS-FR                             & 0.868                               & 26.44                               & 0.078             & 54.28                                & 3.349                             & 0.748                              & 0.844                               & 23.23                               & 0.082             & 61.45                                & 3.337                             & 0.754                               \\
		MLG                               & 0.872                               & 26.19                               & 0.053             & 61.49                                & 3.320                             & 0.721                              & 0.848                               & 24.19                               & 0.078             & 63.22                                & 3.373                             & 0.732                               \\
		TD                                & 0.865                               & 26.23                               & 0.061             & 58.71                                & 3.337                             & 0.714                              & 0.832                               & 23.45                               & 0.072             & 63.76                                & 3.219                             & 0.718                               \\
		\textbf{Ours}                     & \textbf{0.883}                      & \textbf{27.15}                      & \textbf{0.040}    & \textbf{48.03}                       & \textbf{3.543}                    & \textbf{0.773}                     & \textbf{0.856}                      & \textbf{25.86}                      & \textbf{0.068}    & \textbf{56.27}                       & \textbf{3.461}                    & \textbf{0.799}                      \\
		\hline
	\end{tabular}
	\caption{Quantitative results on iPER and ComplexMotion  datasets. We quantitatively assess the performance on two scenarios: Person Reconstruction and Action Imitation. $\uparrow$ indicates the higher is better, $\downarrow$ indicates the lower is better.}
	\label{compare}
\end{table*}

\begin{table*}[ht]\small
	\centering

	\label{ablation}
\begin{tabular}{c|ccc|cc|ccc|cc} 
	\hline
	\multirow{3}{*}{\begin{tabular}[c]{@{}c@{}}\textbf{Methods}\\\end{tabular}} & \multicolumn{5}{c|}{\textbf{ComplexMotion}}                                                                                                           & \multicolumn{5}{c}{\textbf{iPER}}                                                                                                                      \\ 
	\cline{2-11}
	& \multicolumn{3}{c|}{Person Reconstruction}                                                    & \multicolumn{2}{c|}{Action Imitation}                 & \multicolumn{3}{c|}{Person Reconstruction}                                                    & \multicolumn{2}{c}{Action Imitation}                   \\ 
	\cline{2-11}
	& \multicolumn{1}{c|}{SSIM$\uparrow$} & \multicolumn{1}{c|}{PSNR$\uparrow$} & LPIPS$\downarrow$ & \multicolumn{1}{c|}{FID$\downarrow$} & IS$\uparrow$   & \multicolumn{1}{c|}{SSIM$\uparrow$} & \multicolumn{1}{c|}{PSNR$\uparrow$} & LPIPS$\downarrow$ & \multicolumn{1}{c|}{FID$\downarrow$} & IS$\uparrow$    \\ 
	\hline
	\textbf{Our method, Complete}                                               & \textbf{0.883}                      & \textbf{27.15}                      & \textbf{0.040}    & \textbf{48.03}                       & \textbf{3.543} & \textbf{0.856}                      & \textbf{25.86}                      & \textbf{0.068}    & \textbf{56.27}                       & \textbf{3.461}  \\ 
	\hline
	\multicolumn{11}{c}{Ablation Analysis of~\textbf{\textbf{Gromov-Wasserstein loss~}}}                                                                                                                                                                                                                                                                                                         \\ 
	\hline
	L2 loss                                                                     & 0.836                               & 25.06                               & 0.059             & 63.62                                & 3.412          & 0.822                               & 24.10                               & 0.081             & 64.35                                & 3.168           \\ 
	\hline
	\multicolumn{11}{c}{Ablation Analysis of~\textbf{\textbf{Dense Skip Connection}}}                                                                                                                                                                                                                                                                                                            \\ 
	\hline
	r/m dense skip connection                                                   & 0.868                               & 25.28                               & 0.050             & 62.39                                & 3.218          & 0.813                               & 24.21                               & 0.075             & 62.12                                & 3.271           \\ 
	\hline
	\multicolumn{11}{c}{Ablation Analysis of~\textbf{\textbf{Self-Supervised Information}}}                                                                                                                                                                                                                                                                                                      \\ 
	\hline
	Image feature                                                               & 0.703                               & 22.42                               & 0.129             & 83.44                                & 3.215          & 0.634                               & 22.14                               & 0.108             & 99.34                                & 3.019           \\ 
	\hline
	2 face vectors                                                              & 0.728                               & 24.68                               & 0.129             & 78.20                                & 3.108          & 0.719                               & 21.34                               & 0.114             & 89.51                                & 3.167           \\
	3 face vectors                                                              & 0.758                               & 25.62                               & 0.079             & 56.80                                & 3.331          & 807                                 & 22.56                               & 0.089             & 73.33                                & 3.267           \\
	4 face vectors                                                              & 0.784                               & 26.08                               & 0.088             & 59.71                                & 3.304          & 0.753                               & 21.52                               & 0.098             & 75.28                                & 3.261           \\ 
	\hline
	1 candidate face                                                            & 0.732                               & 24.88                               & 0.139             & 75.20                                & 3.158          & 0.689                               & 20.24                               & 0.104             & 88.51                                & 3.067           \\
	2 candidate faces                                                           & 0.793                               & 26.22                               & 0.083             & 60.91                                & 3.371          & 0.746                               & 22.72                               & 0.088             & 75.54                                & 3.321           \\ 
	\hline
	\multicolumn{11}{c}{Ablation Analysis of \textbf{Multiple Local GANs}}                                                                                                                                                                                                                                                                                                                       \\ 
	\hline
	r/m multi-local GAN                                                         & 0.872                               & 26.19                               & 0.053             & 61.49                                & 3.320          & 0.848                               & 24.19                               & 0.078             & 63.22                                & 3.373           \\
	\hline
\end{tabular}
	\caption{Ablation study of different components in our method performed on iPER and ComplexMotion datasets. “r/m X” refers to removing X module in our network. The complete network consistently achieves the best results which are highlighted.}
	\label{ablation}
\end{table*}

The Multi Local GANs polish the details of key body parts and we use critics $D_j$ to present supervision on local details:

\begin{align}
	\begin{split}
		&L_{lr} = \mathbb{E}_{I}[\log D_j({F}^1)] 
		+ \mathbb{E}_I[\log(1-D_j( {\overline{F}}^1+G_j(\overline{{F}}^{\prime 1}))]\\
		&+\sum_{j=2}^5( \mathbb{E}_{I}[\log D_j({F}^j)] 
		+ \mathbb{E}_I[\log(1-D_j( {\overline{F}}^j+G_j(\overline{{F}}^j))]).
	\end{split}
\end{align}

\section{Experiments}
	
	

\subsection{Experimental Settings}

\paragraph{Datasets.} Experiments are conducted on two benchmark datasets, iPER \cite{liu2019liquid} and ComplexMotion. The \textbf{iPER} dataset contains 30 subjects of different shapes, heights and genders. Each subject wears diverse clothes and performs different motions. The whole dataset contains 241,564 frames from 206 videos. 
Although there are a large number of videos in the iPER, the motions in the videos are relatively common. In order to further examine the performance of the methods on complex motion scenes, \emph{such as doing sophisticated and rapid motion}, we collected a benchmark video dataset, \textbf{ComplexMotion}, consisting of 122 in-the-wild videos for more than 50 human subjects. These videos are acquired from various video repositories including TikTok and Youtube. In particular, humans in the dataset wear different clothes and perform complex and rapid movements such as \emph{street dance}, \emph{sports}, and \emph{kung fu}. 

\paragraph{Implementation details.}
During training, all the images are resized to 512 × 512. We utilize OpenPose to detect the 18 human joints in each frame, which form the pose skeleton. We employ the Mask-RCNN to disentangle each video frame into the foreground and background. We train our model with a mini-batch of $10$ for 120 epochs on a Nvidia RTX 2080-Ti GPU. The initial learning rate is set to $1e-4$. We employ the Adam optimizer with $\beta_1$ = 0.9 and $\beta_2$ = 0.999. 

\subsection{Comparisons to State-of-the-art Human Motion Copy Methods}
To evaluate our approach FakeMotion, we compare it with state-of-the-art approaches including Everybody Dance Now (EDN) \cite{chan2019everybody}, PoseWarp \cite{balakrishnan2018synthesizing}, Few-Shot Video2Video (FSV2V) \cite{wang2019few}, Liquid Warping GAN (LWG) \cite{liu2019liquid}, C2F-FWN \cite{wei2020c2f}, SS-FR, MLG, and TD. Specifically, SS-FR, MLG and TD are three variants of FakeMotion, which correspond to removing the \emph{self-supervised face refinement, multiple local GANs}, and \emph{temporal discriminator} from our default method, respectively.

\paragraph{Quantitative Comparisons.} Following the protocol of existing methods \cite{liu2019liquid,huang2021few}, we quantitatively benchmark our method and current approaches on two scenes: \emph{Person Reconstruction} and \emph{Action Imitation}. For \emph{Person Reconstruction}, we perform self-imitation where a person mimics motions from themselves. We adopt Structural Similarity (SSIM) \cite{wang2004image}, Peak Signal to Noise Ratio (PSNR) \cite{korhonen2012peak}, and Learned Perceptual Image Patch Similarity (LPIPS) \cite{zhang2018unreasonable} to evaluate the quality of the generated image.  
For \emph{Action Imitation}, we perform cross-imitation where a person mimics another person's movements. We employ the Inception Score (IS) \cite{salimans2016improved} and Frechet Inception Distance score (FID) \cite{heusel2017gans} as evaluation metrics. To measure the quality of the video instead of image, we also adopt a Temporal Consistency Metric (TCM) \cite{yao2017occlusion} which concerns temporal consistency. The experimental results on ComplexMotion and iPER are reported in Table~\ref{compare}. Our approach consistently outperforms state-of-the-art methods in both \emph{Person Reconstruction} and \emph{Action Imitation}. 

\begin{figure*}[ht]
	\centering
	\includegraphics[width=16cm]{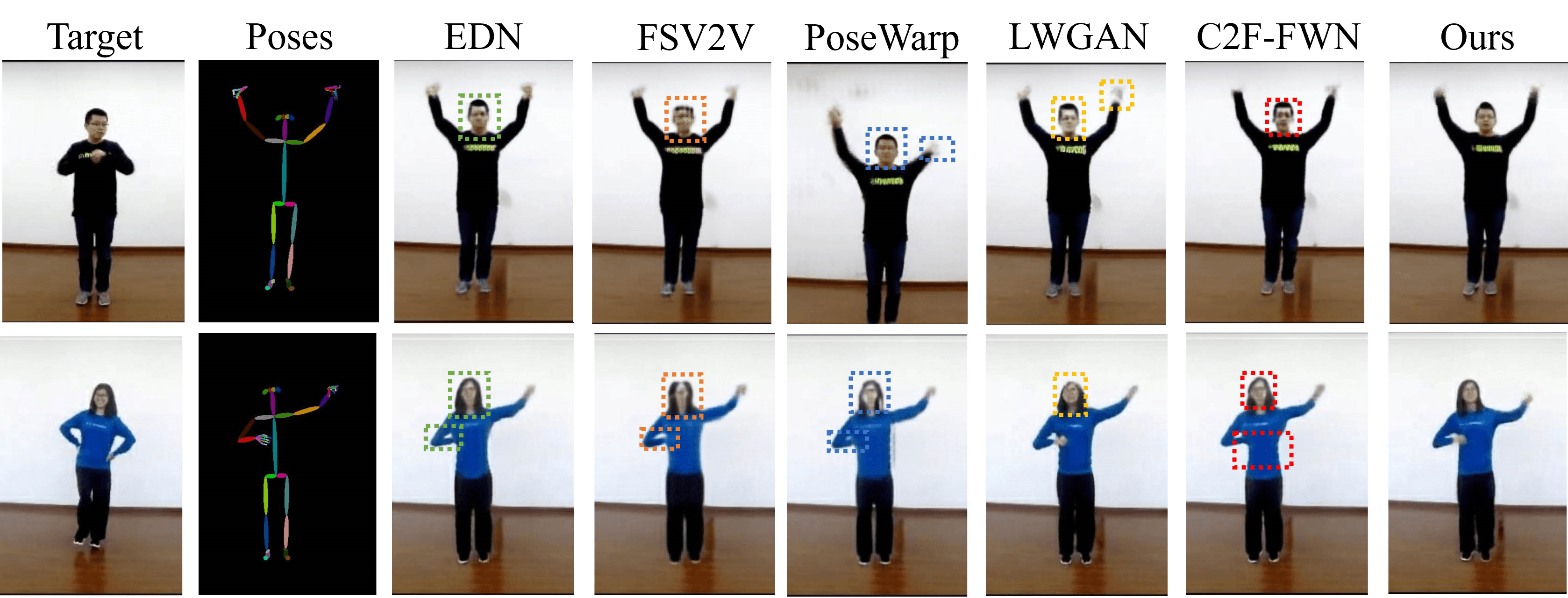}
	\caption{ Visualization of the human video motion copy of our method and state-of-the-art approaches.  Each column from left to right represents target images, source poses, EDN, FSV2V, PoseWarp, LWGAN, C2F-FWN and our method, respectively. Inaccurate appearances are highlighted with dotted rectangles of different colors. Please zoom in to see more details.}
	\label{fig_comparisons}		
\end{figure*}

\paragraph{Qualitative Comparisons.} In addition to quantitative comparisons, we also analyze the generated images visually. Fig.~\ref{fig_comparisons} illustrates visual results of six different methods. We observe that PoseWarp and FSV2V suffer from distorted human body and absent hands. EDN achieves plausible visual results, however, the face of the person is blurred. Warping-based LWGAN and C2F-FWN are able to effectively copy the motion of source person, but fail to generate fine-grained parts such as hairs and clothing. Our method, in contrast,  yields more robust and realistic results on human subjects. We attribute the performance gain to the proposed dense skip connections, Gromov-Wasserstein objective, and the body part GANs. 



\paragraph{Ablation Study.}
Extensive ablation experiments are also conducted  to investigate the effects of various components. 
We first investigate the effect of removing \textit{Gromov-Wasserstein loss} in the generator. Specifically, we adopt a traditional L2 loss instead of the Gromov-Wasserstein loss. From Table~\ref{ablation}, we observe that SSIM and PSNR fall from 0.883 and 27.15 to 0.836 and 25.06, respectively. These performance deteriorations  upon removal of the Gromov-Wasserstein loss suggest the efficacy of the proposed loss.  In addition to Gromov-Wasserstein loss, we also investigate the impact of removing \textit{dense skip connections} in the generator. From Table~\ref{ablation}, we observe clear performance dips upon removal of the dense skip connection component. This is in line with our intuition that integrating and fusing feature representations from multiple encoder layers are beneficial. 



To evaluate the  \emph{self-supervised face refinement} module, we consider 1) selecting similar real face images based on facial image features instead of face orientations, 2) using a different number of face vectors, and 3) employing a different number of similar real images. The results of these ablation experiments are reported in Table~\ref{ablation}.
We observe significant performance drops of SSIM and PSNR upon adopting the first setting. This suggests the effectiveness of using face orientations for similar face searching, which avoids being interfered by unimportant features such as \emph{color} and \emph{eye shape}.  Furthermore, we observe that the quality of the generated image gradually increases with the growing of the number of face vectors. Similarly, the increase in the number of used similar real images also leads to better results. 



Finally, we also try removing the \emph{multiple local GANs} to evaluate their contributions. Upon the removal,  the FID dramatically increases from $48.03$ to $61.49$ (the smaller the FID value, the better the image quality). This 
highlights the importance of local refinement. In particular, the residual images, which are produced by the multiple local GANs, disclose the difference in fine-scale texture details between the generated local body parts and the ground-truth, facilitating the generation of more lifelike local images.

\section{Related Work}



\paragraph{Pose-guided approaches.} The researches learn mapping functions between a pose and its corresponding image \cite{ren2020deep,zhang2021pise,kappel2021high}. \cite{ma2017pose,wang2018video} seek high-resolution video2video synthesis by adopting GAN or Two scale-GAN networks. \cite{yang2020transmomo} performs human video motion transfer in an unsupervised manner, taking advantage of the independence of three orthogonal factors, \emph{motion, skeleton, and view}. However, these methods fail to take into account the importance of maintaining facial details  and the end effectors (\emph{i.e.} hand and feet) in motion transfer. \cite{chan2019everybody} introduces a face enhancement module that adopts a traditional GAN architecture. 

\paragraph{Warpping-based approaches.} The studies aim to warp the appearance of the target person to copy motions from another person \cite{balakrishnan2018synthesizing,dong2018soft,liu2021liquid,siarohin2021motion}.  
\cite{liu2019liquid}  disentangles the human image into action and appearance, and achieves motion copy using a Warping GAN which warps the image of the target person performing standard actions. \cite{wei2020c2f} extracts the motion manifold of the target human and accordingly refines the detailed motion. However, these methods are mesh based, and have inherent difficulties in dealing with irregular shapes and clothing, and also suffer major performance deterioration during rapid motions (unless we have an unlimited number of meshes).


\section{Conclusion}
\label{conclusion}
In this work, we propose a novel fake video generation framework termed \emph{FakeMotion}, which is equipped with dense skip connections, Gromov-Wasserstein GANs, and local enhancement modules. A particular highlight of our method is the Gromov-Wasserstein loss which enables a robust and stable generator training.
We further propose a self-supervised face refinement module that resorts to real faces with similar orientations to polish facial details (textures), and multiple local GANs which enhance local details of body parts.  Extensive experiments show that our method achieves notable improvements and is able to handle rapid-motion and complex scenes.

\section*{Acknowledgements}
This research is supported by the National Key R\&D Program of China (Grant no.2018YFB1404102), the National Natural Science Foundation of China (No. 61902348), the Key R\&D Program of Zhejiang Province (No. 2021C01104), and the Postgraduate Research Innovation Fund Project of Zhejiang Gongshang University.

\bibliographystyle{named}
\bibliography{ijcai22}

\end{document}